\def\year{2022}\relax
\documentclass[letterpaper]{article} 
\usepackage{aaai22}  
\usepackage{times}  
\usepackage{helvet}  
\usepackage{courier}  
\usepackage[hyphens]{url}  
\usepackage{graphicx} 
\usepackage{natbib}  
\usepackage{caption} 
\DeclareCaptionStyle{ruled}{labelfont=normalfont,labelsep=colon,strut=off} 
\usepackage{algorithm}
\usepackage{algorithmic}
\usepackage{color}

\usepackage{newfloat}
\usepackage{listings}
\floatstyle{ruled}
\newfloat{listing}{tb}{lst}{}
\floatname{listing}{Listing}
\usepackage{graphicx}
\usepackage{subfigure}
\usepackage{multirow}
\usepackage{tcolorbox}

\title{STFL: A Spatial-Temporal Federated Learning Framework for Graph Neural Networks}
\author {
    Guannan Lou\textsuperscript{\rm 1}, 
    Yuze Liu\textsuperscript{\rm 2}, 
    Tiehua Zhang\textsuperscript{\rm 2}\thanks{Corresponding author.}, 
    Xi Zheng\textsuperscript{\rm 1} 
}
\affiliations {
    \textsuperscript{\rm 1} Macquarie University\\
    \textsuperscript{\rm 2} Ant Group\\
    lougnroy@gmail.com, liuyuze.liuyuze@antgroup.com, zhangtiehua.zth@antgroup.com, james.zheng@mq.edu.au
}

\usepackage{bibentry}

\begin{document}

\maketitle

\begin{abstract}
We present a spatial-temporal federated learning framework for graph neural networks, namely STFL. The framework explores the underlying correlation of the input spatial-temporal data and transform it to both node features and adjacency matrix. The federated learning setting in the framework ensures data privacy while achieving a good model generalization. Experiments results on the sleep stage dataset, ISRUC\_S3, illustrate the effectiveness of STFL on graph prediction tasks. 
\end{abstract}

\section{Introduction}

Graph Neural Network (GNN) has emerged recently owing to its ability to learn vector representations from complex graph data. The GNN has now been widely used in applications such as social network recommendation \cite{wu2021fedgnn, he2019cascade}, traffic flow prediction~\cite{wang2020traffic, cui2019traffic}, action recognition~\cite{yan2018spatial} and medical diagnosis~\cite{rong2020self, sun2020graph}. In addition to only using GNN models on learning the graph representations from different graph data, one critical question is how to generalize the GNN models even when the training data is insufficient, regardless of the graph or non-graph structures. This scenario applies to almost all cases where the data privacy is the major concern, and the model can only be trained to match the distribution of the local dataset. For instance, hospitals and drug research institutions rarely publish patient data and clinical data due to its sensitivity, and the trained GNN models thus fail to produce the proper node/graph representations for data that are not present in the training set.

To this end, Federated Learning (FL) is designated to solve such problems. It is a distributed model training method that can be used to deal with data isolation between different data sources~\cite{kairouz2019advances}. In FL, all clients participate in training only using local data while not exposing it to others. By integrating all client models weights or gradients, the FL trained model will thus have a higher generalization ability.

Currently, there are only a few works focusing on implementing FL on GNNs~\cite{he2021fedgraphnn, wu2021fedgnn}. FedGraphNN~\cite{he2021fedgraphnn} is an open FL benchmark system for training GNN, which demonstrates good performance on 36 graph-structured datasets in domains like social networks and recommendation systems. Similarly, FedGNN~\cite{wu2021fedgnn} implement the GNN under an FL framework to ensure data privacy in recommendation tasks. However, one common attribute of existing FL on GNNs works aims at training on well-established graph-structured data. 
In the real-world scenario, not every dataset has the built-in graph structure, and there is a colossal amount of spatial-temporal data, presenting a intriguing research question in terms of how to input such data into GNNs. Therefore, it is of great significance to design such an end-to-end FL-GNN training framework.

In this paper, we propose a novel end-to-end spatial-temporal federated learning framework for graph neural networks (STFL), which can automatically transform time series data into graph-structured data, and train GNNs collectively to ensure good data privacy and model generalization. 
We break our contributions into the following parts: 1. we first implement the graph generator to handle spatial-temporal data, including both feature extraction and node correlation exploration; 2. we integrate the graph generator into the proposed STFL and design an end-to-end federated learning framework for spatial-temporal GNNs on graph-level classification tasks;  3. we perform extensive experiments on real-world sleeping dataset: ISRUC\_S3; 4. we publish the source code of STFL on Github~\footnote{https://github.com/JW9MsjwjnpdRLFw/TSFL}.

\begin{figure*}[t]
\centering
\includegraphics[width=0.8\linewidth]{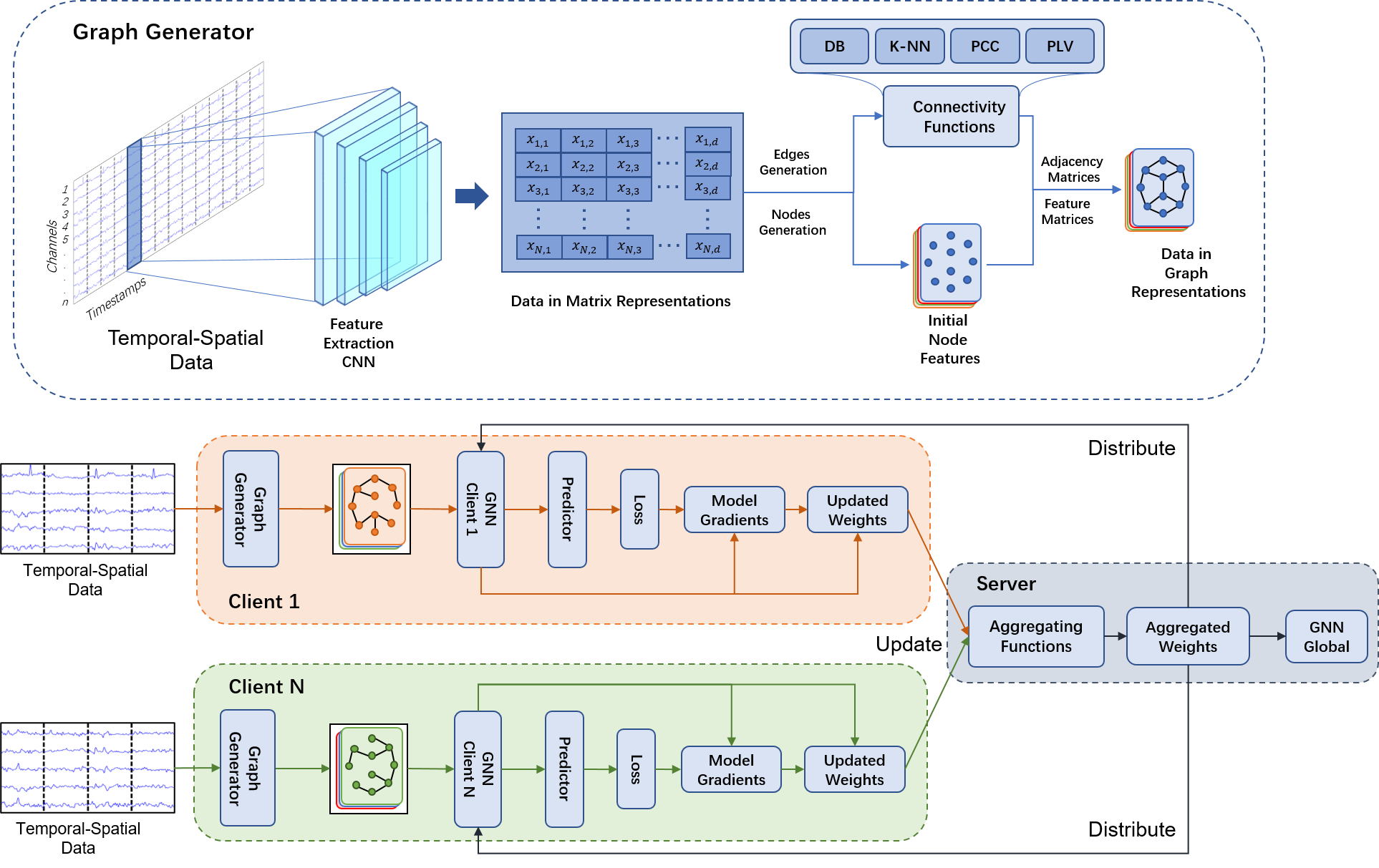} 
\caption{The framework of STFL.}

\label{fig:process}
\end{figure*}

\section{Related Work}

\subsubsection{Spatial-Temporal Graph Neural Networks}

Building graph-structured data from the spatial-temporal sequence has been a challenging task, and such data is considered beneficial to traffic flow prediction, action recognition and brainwave recognition tasks. There are generally two main hurdles when transforming the spatial-temporal sequence to GNN-required graph input: 1. to uncover the spatial correlations of the data streams to generate the adjacency matrix; 2. to extract the node features from the temporal values.
Spatial-Temporal Graph Neural Networks~\cite{jain2016structural, yu2017spatio, seo2018structured, yan2018spatial} are proposed to solve these issues. Graph Convolutional Recurrent Network (GCRN)~\cite{seo2018structured} combines the LSTM network with ChebNet~\cite{defferrard2016convolutional} to handle spatial-temporal data. Structural-RNN~\cite{jain2016structural} uses node-level and edge-level RNN to represent spatial relations of data. Alternatively, CNN could be used to embed temporal relationships to solve the hidden issues of gradient explosion/vanishing. For example, ST-GCN~\cite{yan2018spatial} uses the partition graph convolution layer to extract spatial information, and use the one-dimension convolution layer to extract temporal relationships. Similarly, CGCN~\cite{yu2017spatio} combine an one-dimension convolution layer with a ChebNet or GCN layer, to handle spatial-temporal data.
Comparing with CGCN and ST-GCN, STFL uses the same strategy, which uses CNN to capture the spatial dependency and GCN to model the temporal dependency. Moreover, STFL not only illustrates how to handle Spatial-Temporal data, but also explores the federated learning process following that.

\subsubsection{Federated Learning on Graph Neural Network}

In order to solve the issue of lacking data and preserve local data privacy, recent works focus on training GNNs under federated learning settings, which can be broadly divided into three categories~\cite{zhang2021federated}: inter-graph FL, intra-graph FL and graph-structured FL. In inter-graph FL, each client is assigned with the full graph and the global GNN performs graph-level tasks like medical diagnosis~\cite{rong2020self, sun2020graph}. In intra-graph FL, each client only possesses the local graph, which is a part of the whole graph, and the global model performs both on node-level or edge-level tasks. Social network recommendation is the typical task of this type~\cite{wu2021fedgnn, he2019cascade}. For graph-structured FL, each client behaves as one node in the graph, and the topology of clients are considered as the edges in the graph. The traffic flow prediction \cite{meng2021cross} is the typical application, where monitoring devices are clients in this case, and their geographic positions and distance are used to construct the graph.  

In this work, STFL belongs to inter-graph FL. 
Each client in STFL is assigned with a complete GNN generated automatically from the underlying spatial-temporal data,
with the goal of training the global model for graph-level prediction tasks. The STFL takes the end-to-end training approach, encompassing the graph construction, federated learning for GNN, and label prediction. We believe STFL is the first of its kind in end-to-end intra-graph FL framework for GNN on spatial-temporal data.

\section{Methodology}
We elaborate on the proposed framework in this section and formalize the graph generation, graph neural network and federated learning, respectively.

\subsection{Graph Generation}
As mentioned in early sections, we consider the spatial-temporal sequence as the raw input. Define a multi-variate series 
$\emph{S} = \left[\emph{s}_{1},...,\emph{s}_{T}\right]^{\mathsf{T}}\in\mathcal{R}^{T\times D}$
as a temporal series set, where there are total $T$ timestamps, each of which has the $\emph{s}_{i}\in\mathcal{R}^{D}$ dimensional signal frequency. Since there is no node concept in the spatial-temporal data, we thus take advantage of spatial channels and treat it as nodes, meaning if there are $N$ channels, there will be $N$ nodes in the transformed graph data structure. Say each channel has a temporal series set $S$, the spatial-temporal sequence with full channels is then denoted as $\emph{C}_{raw}\in\mathcal{R}^{N\times T\times D}$. The structure of the spatial-temporal data can be seen in the graph generator part in Fig.~\ref{fig:process}. Following that, a CNN-based feature extraction net~\cite{9530406} is used to convert the raw spatial-temporal data into the feature matrix representations. 
The detailed model structure of feature extraction net can be found in Table~\ref{tab:FeatureNet}.
The output of feature extraction net is $\emph{C}_{re}\in\mathcal{R}^{T\times N\times d}$, where $d$ denotes the dimension of refined features. A snapshot of the $\emph{C}_{re}$ is represented as $\emph{X}_{T}\in\mathcal{R}^{N\times d}$  and can also be seen in Fig.~\ref{fig:process}'s graph generator part.

\begin{table}[]
\resizebox{\linewidth}{!}{
\begin{tabular}{l|rrr|rrr}
\hline \hline
            & Output dim & Filters & Activation & Output dim & Filters & Activation \\ \hline
Input       & 3000×1     &         &            &            &         &            \\ \hline
Conv1D+BN   & 492×32     & 32      & ReLU       & 53×64      & 64      & ReLU       \\
MaxPool1D   & 30×32      & -       & -          & 6×64       & -       & -          \\
Dropout     & 30×32      & -       & -          & 6×64       & -       & -          \\
Conv1D+BN   & 30×64      & 64      & ReLU       & 6×64       & 64      & ReLU       \\
Conv1D+BN   & 30×64      & 64      & ReLU       & 6×64       & 64      & ReLU       \\
Conv1D+BN   & 30×64      & 64      & ReLU       & 6×64       & 64      & ReLU       \\
MaxPool1D   & 3×64       & -       & -          & 1×64       & -       & -          \\
Flatten     & 192        & -       & -          & 64         & -       & -          \\ \hline
Concatenate & 256        &         &            &            &         &            \\ \hline
\end{tabular}}
\caption{Structure of feature extraction net, which is a two branch CNN model. One branch (left-handed) has a smaller convolution kernel to capture temporal information, while another (right-handed) has larger convolution kernels to capture frequency information. Concatenation from the results of both two branch generates the node feature matrix. \cite{9530406}}
\label{tab:FeatureNet}
\end{table}

After obtaining the refined feature matrix $\emph{C}_{re}$, the correlation between channels (nodes) needs to be uncovered. It is natural to treat $\emph{X}_{T}\in\mathcal{R}^{N\times d}$ as node feature matrix at this point and retrieve the underlying correlation among them. Henceforth, we define node correlation functions, which takes the node feature matrix as the input and outputs the required adjacency matrix $\mathcal{A}_{T} \in \mathcal{R}^{N\times N}$:

\begin{equation}
    \mathcal{A}_{T} = \emph{Corr}\left(\emph{X}_{T}\right)
\end{equation}

where $\emph{Corr}\left(\cdot\right)$ computes correlations or dependencies of each channels (nodes) on the basis of $\emph{X}_{T}$. There are several choices of node correlation functions, such as 
Pearson correlation function \cite{pearson1903laws} or phase locking value function \cite{aydore2013note}, etc. We will take a close look at each one of them in the experiment section.

The graph $\mathcal{G}_{T}$ can then be assembled using both $\mathcal{A}_{T}$ and $\emph{X}_{T}$. 
\subsection{Graph Neural Network}
Along the time dimension, We obtain $\left\{\mathcal{G}_{1},...,\mathcal{G}_{T}\right\}$ as the whole graph datasets, representing the generated graph data at each timestamp, and we use $\left\{\emph{y}_{1},...,\emph{y}_{T}\right\}$ to correspond to graph labels. We formulate the graph prediction task here in which the output of the graph generator expects to be predicted correctly. For notation simplicity, we use $\mathcal{V}_{T}$ to denote the node set in each $\mathcal{G}_{T}$, and the number of node $\|V\|$ is essentially same as row number in node feature matrix $\emph{X}_{T}$. For each $v \in\mathcal{V}$,
respective node feature is written as $\emph{x}_{v}\in\mathcal{R}^{d}$.
We use $ne\left[v\right]$ to denote the neighborhood of node $v$, the correlation value of which can be retrieved from adjacency matrix $\mathcal{A}$.

We then formulate the message passing and readout phase of the GNN. Let
$\emph{h}_{v}^{l}$ represent the node embedding in layer $l$ and $\emph{h}_{v}^{0} = \emph{x}_{v}$. The message passing of node $v$ from layer $l$ to layer $l+1$ can be formalized as:

\begin{equation}
    \emph{h}^{l+1}_{v} = \sigma\left(\emph{W}^{l+1}\cdot\emph{MEAN}\left\{\emph{h}^{l}_{u}:\forall u\in ne\left[v\right]\cup\left\{v\right\}\right\}\right)
\end{equation}

Here, $\emph{W}^{l+1}$ is learnable transformation matrix at layer $l+1$, $\sigma\left(\cdot\right)$ denotes an activation function like $ReLU$.
GNN updates node embedding $\emph{h}^{l+1}_{v}$ by aggregating the all neighbor representations and itself.

To get representation of the entire graph $\emph{h}^{\mathcal{G}}$ after $L$ message passing layers, GNN perfroms the readout operation to derive the final graph representations from node embeddings, which can be formulated as follows:

\begin{equation}
    \emph{h}^{\mathcal{G}} = \emph{Readout}\left(\left\{\emph{h}^{L}_{u},u\in\mathcal{V}\right\}\right)
\end{equation}

$\emph{Readout}\left(\cdot\right)$ is a permutation invariant operation, which can either be simple mean function or more sophisticated graph-level pooling function like MLP. 

In full supervised setting, We use a shallow neural network  to learn the map between graph embedding $\emph{h}^{\mathcal{G}}$ and label space $\mathcal{Y}$. $\sigma\left(\cdot\right)$ is a non-linear transformation (Sigmoid function in this case), which can be generalized as:

\begin{equation}
    \emph{z}^{\mathcal{G}} = \sigma\left(\emph{W}\cdot \emph{h}^{\mathcal{G}}\right)
\end{equation}

Furthermore, we utilize graph-based binary cross entropy loss function to calculate loss $\mathcal{L}$ in the supervised setting. The loss function is formulated as:


\begin{equation}
    \mathcal{L}\left(\emph{z}^{\mathcal{G}},\emph{y}^{\mathcal{G}}\right) = -\sum\left[\emph{y}^{\mathcal{G}}\cdot\log\emph{z}^{\mathcal{G}}+\left(1-\emph{y}^{\mathcal{G}}\right)\cdot\log\left(1-\emph{z}^{\mathcal{G}}\right)\right]
\end{equation}
where $\emph{z}^{\mathcal{G}}$ is the predicted label of current graph and $\emph{y}^{\mathcal{G}}$ is the ground truth.

\begin{figure*}[t]
\centering
\includegraphics[width=0.95\linewidth]{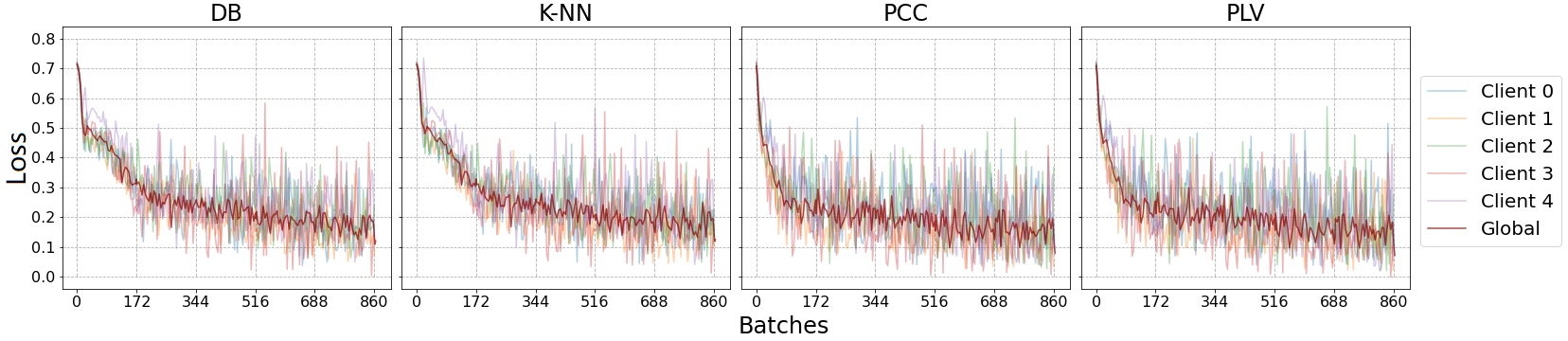} 
\caption{Training and testing loss per batch of Fed-GCN on ISRUC\_S3 with different node correlation functions. The vertical dashed line separates five epochs.}
\label{fig:GCN_loss}
\vspace{0.2in}
\end{figure*}


\begin{figure*}[t]
\centering
\includegraphics[width=0.8\linewidth]{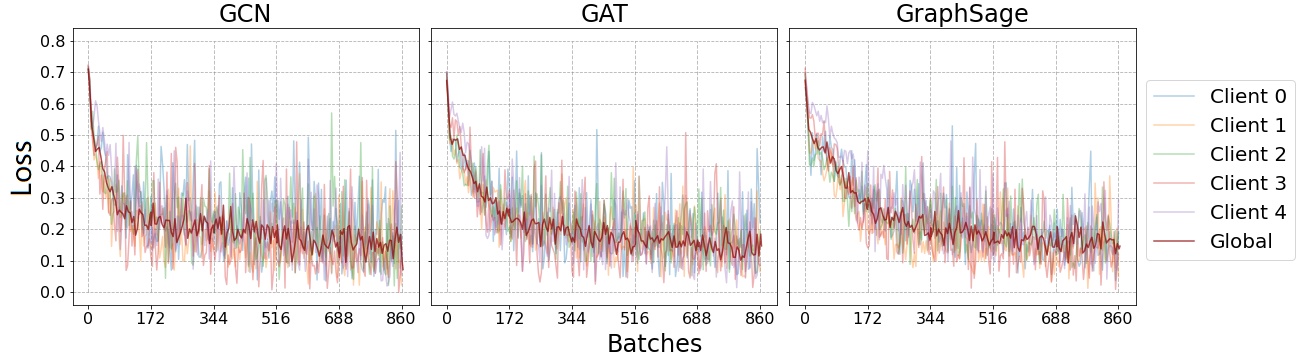} 
  \caption{Training and testing loss of three federated graph models on ISRUC\_S3 with PLV.}
	\label{fig:PLV_loss}
	\vspace{0.2in}
\end{figure*}

\subsection{Federated Learning}
STFL trains GNNs from different clients under federated learning settings. As shown in Figure~\ref{fig:process}, STFL contains a main central server $S$ and $n$ clients $C$. Each client $c_i\in C$ deploys one GNN, which learns from local graph data $\mathcal{G}_{i}$ of the client, and upload weights of the GNN $\mathcal{W}_{i}$ to the central server. The central server receives weights from all clients, update weight $\mathcal{W}_{S}$ of the global GNN  model, and distributes the updated weights back to each client. In this work, we choose the $FedAvg$~\cite{mcmahan2017communication} as the aggregating function, which averages the weight from each client to generate the weight of global GNN on the server. 

\begin{equation}
    \mathcal{W}_{S} = FedAvg(\mathcal{W}_{C}) = \frac{1}{n}\sum_{i=1}^{n}\mathcal{W}_{c_i}
\end{equation}

\section{Experiment}
In this section, we describe the details of our experiment, aiming to answer he following research questions: 
\begin{itemize}
    \item \textbf{RQ1:} what is the influence on using different node correlation functions?
    \item \textbf{RQ2:} how well STFL performs on the real-world dataset?
    \item \textbf{RQ3:} which widely used graph networks achieves better results?
\end{itemize}

\subsection{Experimental Settings}

\subsubsection{Dataset}
In our experiments, ISRUC\_S3~\cite{khalighi2016isruc} is used as the benchmark dataset. ISRUC\_S3 collects polysomnography (PSG) recordings in 10 channels from 10 healthy subjects (i.e., sleep experiment participants). These PSG recordings are labeled with five different sleep stages according to the AASM standard~\cite{jia2020graphsleepnet}, including Wake, N1, N2, N3 and REM. As mentioned in the preceding section, we adopt the CNN-based feature extraction net~\cite{9530406} to generate the initial node features. To generate the adjacency matrix, four different node correlation functions are implemented and discussed separately. To evaluate the effectiveness of STFL, we follow the non-iid data setup~\cite{zhang2020achieving} and split the different sleep stages to clients to verify the effectiveness of our proposed framework.

\subsubsection{Node Correlation Functions}
To study the influence of the generated adjacency matrix on the proposed framework, we experiment on four different node correlation functions, each of which adopts different strategies to quantify the node correlation. These four functions are Distance-Based (DB), $K$-Nearest Neighbor ($K$-NN), Pearson Correlation Coefficient (PCC) and Phase Locking Value (PLV). The different node correlation functions are described as below:
\begin{itemize}
    \item DB is the an euclidean distance function to measure actual spatial distance between pairs of electrodes.
    \item $K$-NN~\cite{jiang2013graph} generates the adjacency matrix that only selects the k nearest neighbor of each node to represent the node correlation of the graph.
    \item PCC~\cite{pearson1903laws} is known as the Pearson correlation function to measure similarity between each pairs of nodes.
    \item PLV~\cite{aydore2013note} is a time-dependent node correlation function to measure signals of each pairs of nodes.
\end{itemize}

\subsubsection{Federated Setup and Hyperparameters}
The overarching goal is to investigate the feasibility of the end-to-end federated graph learning framework, we choose the classic federated average~\cite{mcmahan2016federated} as the weights aggregator in STFL. For GNN models, we evaluate the performance of three widely used GNN models under the proposed framework on the basis of graph classification tasks, including GCN~\cite{kipf2016semi}, GAT~\cite{velivckovic2017graph}, and GraphSage~\cite{hamilton2017inductive}.

When it comes to the experiment settings, we define five training clients, each of which is delicately assigned with non-overlapping labels. 
The test data is randomly sampled with the ratio of 0.25 to the whole dataset. 
The data from the rest is first sorted by labels, then divided into 15 partitions, and each client is assigned with 3 partitions with 3 labels.
We use Adam~\cite{kingma2014adam} as the optimizer with the learning rate of 0.015. The dropout~\cite{srivastava2014dropout} ratio is set to be 0.3. Models are trained for 5 epochs, and the batch size of each is 8. We use F1 score and accuracy as evaluation metrics for this experiment.  

\subsection{Performance Comparison and Analysis}

\subsubsection{The Study on Node Correlation Functions (\textbf{RQ1})}


\begin{table}[t]
 \centering
 \resizebox{0.9\linewidth}{!}{
\begin{tabular}{l|cccc}
\hline\hline
\multicolumn{1}{c|}{\multirow{2}{*}{\textbf{Methods}}} & \multicolumn{4}{c}{\textbf{F1}}                              \\ \cline{2-5} 
\multicolumn{1}{c|}{}                                  & \textbf{DB} & \textbf{$K$-NN} & \textbf{PCC}  & \textbf{PLV}   \\ \hline
Fed-GCN                                                & 0.819       & 0.814           & 0.832         & \textbf{0.841} \\
Fed-GAT                                                & 0.838       & 0.839           & 0.849         & \textbf{0.852} \\
Fed-GraphSage                                          & 0.848       & \textbf{0.854}  & 0.848         & 0.848          \\ \hline\hline
\end{tabular}}
\caption{F1 score of GNN models under federated settings on ISRUC\_S3 with different node correlation functions.}
\label{tab:F1}
\end{table}

To evaluate the effectiveness of four node correlation functions, we compare the influence of each on GCN under federated settings, because GCN has the simplest structure among three GNN models. As shown in Figure~\ref{fig:GCN_loss}, PCC and PLV work well under federated settings with faster converging rate, especially in the first two epochs. Furthermore, comparing with other node correlation functions, as shown in Table~\ref{tab:F1}, the F1 score of PLV on three federated models is averagely the highest, followed by PCC, and DB is the worst.
This may be due to the pooling layers in CNN model (feature extraction net), which looks at small time window of the input sequence, from which correct correlations for each pair of nodes can be extracted using PLV.

\subsubsection{The Performance of STFL \textbf{(RQ2)}}


\begin{table}[t]
\centering
\resizebox{0.9\linewidth}{!}{
\begin{tabular}{cc|ccc}
\hline\hline
\multicolumn{2}{c|}{}                                                 & \textbf{GCN}   & \textbf{GAT}   & \textbf{GraphSage} \\ \hline
\multicolumn{1}{c|}{\multirow{2}{*}{\textbf{Fed (PLV)}}} & \textbf{F1}  & 0.841          & 0.852          & 0.848              \\
\multicolumn{1}{c|}{}                                  & \textbf{ACC} & 0.856          & 0.860          & 0.857              \\ \hline
\multicolumn{1}{c|}{\multirow{2}{*}{\textbf{Cen (PLV)}}} & \textbf{F1}  & 0.807          & 0.792          & 0.815     \\
\multicolumn{1}{c|}{}                                  & \textbf{ACC} & 0.818          & 0.804          & 0.834     \\ \hline\hline
\end{tabular}}
\caption{F1 score and accuracy of federated models and centralized models on ISRUC\_S3, using PLV as the node correlation function.}
\label{tab:PLV_F1_ACC}
\end{table}

To evaluate the effectiveness of STFL, we test its performance from different perspectives.
In our experiments, we first evaluate the federated graph models on ISRUC\_S3 with PLV, because PLV performs best among four node correlation functions which is discussed in the RQ1. As shown in Table~\ref{tab:PLV_F1_ACC}, under STFL, all three GNN models produce reasonable results. In particular, under federated settings, GAT achieves the highest F1 score and accuracy on the PLV, and GraphSage comes second. 

Furthermore, we check the result of centralized models of these three graph networks, and the result is also illustrated in Table~\ref{tab:PLV_F1_ACC}. In this part, the hyperparameters remain the same with federated experiments. For the splits of data, the testing data is same with that in federated learning experiments. The training data is randomly sampled from the data aggregated from all clients, and the size of the training data is kept same as that of one client. 
For all GNNs under centralized settings, GraphSage achieves the highest F1 scores and accuracy, followed by GCN. Moreover, all models trained under federated setting achieves a superior results (both F1 score and accuracy) compared to that in centralized setting. This shows that model trained under STFL successfully generalizes the data distribution under non-iid setup. Another finding is that the best GNN model in the centralized setting is not necessarily the best in federated setting.

\subsubsection{The Analysis on Graph Neural Networks (\textbf{RQ3})}


\begin{table}[t]
\centering
\resizebox{0.95\linewidth}{!}{
\begin{tabular}{l|ccccc}
\hline\hline
\multicolumn{1}{c|}{\multirow{2}{*}{\textbf{Methods}}} & \multicolumn{5}{c}{\textbf{PLV}}                                                      \\ \cline{2-6} 
\multicolumn{1}{c|}{}                                  & \textbf{Wake}  & \textbf{N1}     & \textbf{N2}     & \textbf{N3}     & \textbf{REM}   \\ \hline
Fed-GCN                                                & \textbf{0.924} & 0.719           & 0.843           & \textbf{0.916}  & 0.802          \\
Fed-GAT                                                & 0.895          & 0.747           & \textbf{0.859}  & 0.909           & \textbf{0.852} \\
Fed-GraphSage                                          & 0.912          & \textbf{0.751}  & 0.838           & 0.909           & 0.831          \\ \hline\hline
\end{tabular}}
\caption{F1 score for each class of federated models on ISRUC\_S3 with PLV.}
\label{tab:Class_F1}
\end{table}

To investigate the best match of GNNs with STFL, three GNNs are tested on ISRUC\_S3 with PLV under the federated framework, as PLV is observed to achieve best the results among all node correlation functions, the details of which are analyzed in RQ1. As shown in Fig~\ref{fig:PLV_loss}, the GCN converges the fastest, but is much more volatile than the other two. We also discovered that GraphSage converges slowest in the first epoch but achieves a stable loss decrease in the test stage. It is also found that all three models eventually converge to the same loss, fluctuating at around 0.15. Furthermore, we evaluate the F1 score on each class using PLV. Table~\ref{tab:Class_F1} illustrates that, for REM, GraphSage performs the best, while GCN receives the highest score for other four classes.

It is interesting to see the training loss of three models are fluctuated in a large scale, especially in the last three epochs. It may because that the federated framework distributes the global model to each client in each training batch. At the late stage of training, each client cannot fit its own data well in the generalized global model, especially for those models that are prone to overfitting.

\section{Conclusion and Future Work}

Experiment results not only illustrate the effectiveness of STFL in dealing with spatial-temporal data, but also in training GNN in a collective manner. It is interesting to investigate the extendibility of STFL when using versatile data structure, and how well STFL handles graph-level and node-level tasks in the same time. Furthermore, we also need to evaluate different aggregation functions in the federated setting other than FedAvg.

\clearpage
\bibliography{aaai22.bib}

\end{document}